\pdfoutput=1

\documentclass[11pt]{article}

\usepackage[final]{acl}

\usepackage{times}
\usepackage{latexsym}
\usepackage{amsmath}
\usepackage{booktabs}
\usepackage{amssymb}
\usepackage{geometry}
\usepackage{multirow}
\usepackage{xcolor}
\usepackage[T1]{fontenc}
\usepackage[utf8]{inputenc}
\usepackage{microtype}
\usepackage{inconsolata}
\usepackage{graphicx}
\title{Elevating Legal LLM Responses: Harnessing Trainable Logical Structures and Semantic Knowledge with Legal Reasoning}

\author{
 \textbf{Rujing Yao\textsuperscript{1}},
 \textbf{Yang Wu\textsuperscript{2}},
 \textbf{Chenghao Wang\textsuperscript{3}},
 \textbf{Jingwei Xiong\textsuperscript{4}},
 \textbf{Fang Wang\textsuperscript{1}},
 \textbf{Xiaozhong Liu\textsuperscript{2}$^{*}$}
\\
 \textsuperscript{1}Nankai University, Tianjin, China \\
 \textsuperscript{2}Worcester Polytechnic Institute, USA \\
 \textsuperscript{3}Fayuan Technology Co., Ltd., Hangzhou, China \\ 
 \textsuperscript{4}University of California, Davis, USA
\\
\small\texttt{rjyao@mail.nankai.edu.cn, ywu19@wpi.edu, chenghao.wch@gmail.com, jwxxiong@ucdavis.edu,}\\
\small\texttt{wangfangnk@nankai.edu.cn, xliu14@wpi.edu}}

\begin{document}
\maketitle
\begin{abstract}
Large Language Models (LLMs) have achieved impressive results across numerous domains, yet they experience notable deficiencies in legal question-answering tasks. LLMs often generate generalized responses that lack the logical specificity required for expert legal advice and are prone to hallucination, providing answers that appear correct but are unreliable. Retrieval-Augmented Generation (RAG) techniques offer partial solutions to address this challenge, but existing approaches typically focus only on semantic similarity, neglecting the logical structure essential to legal reasoning. In this paper, we propose the Logical-Semantic Integration Model (LSIM), a novel supervised framework that bridges semantic and logical coherence. LSIM comprises three components: reinforcement learning predicts a structured fact-rule chain for each question, a trainable Deep Structured Semantic Model (DSSM) retrieves the most relevant candidate questions by integrating semantic and logical features, and in-context learning generates the final answer using the retrieved content. Our experiments on a real-world legal QA dataset-validated through both automated metrics and human evaluation-demonstrate that LSIM significantly enhances accuracy and reliability compared to existing methods. 
\end{abstract}

\renewcommand{\thefootnote}{\fnsymbol{footnote}}
\footnotetext[1]{Corresponding author.}
\renewcommand{\thefootnote}{\arabic{footnote}}

\section{Introduction}

\begin{figure}[t]
\setlength{\belowcaptionskip}{-0.3cm}
    \centering
    \includegraphics[width=\linewidth]{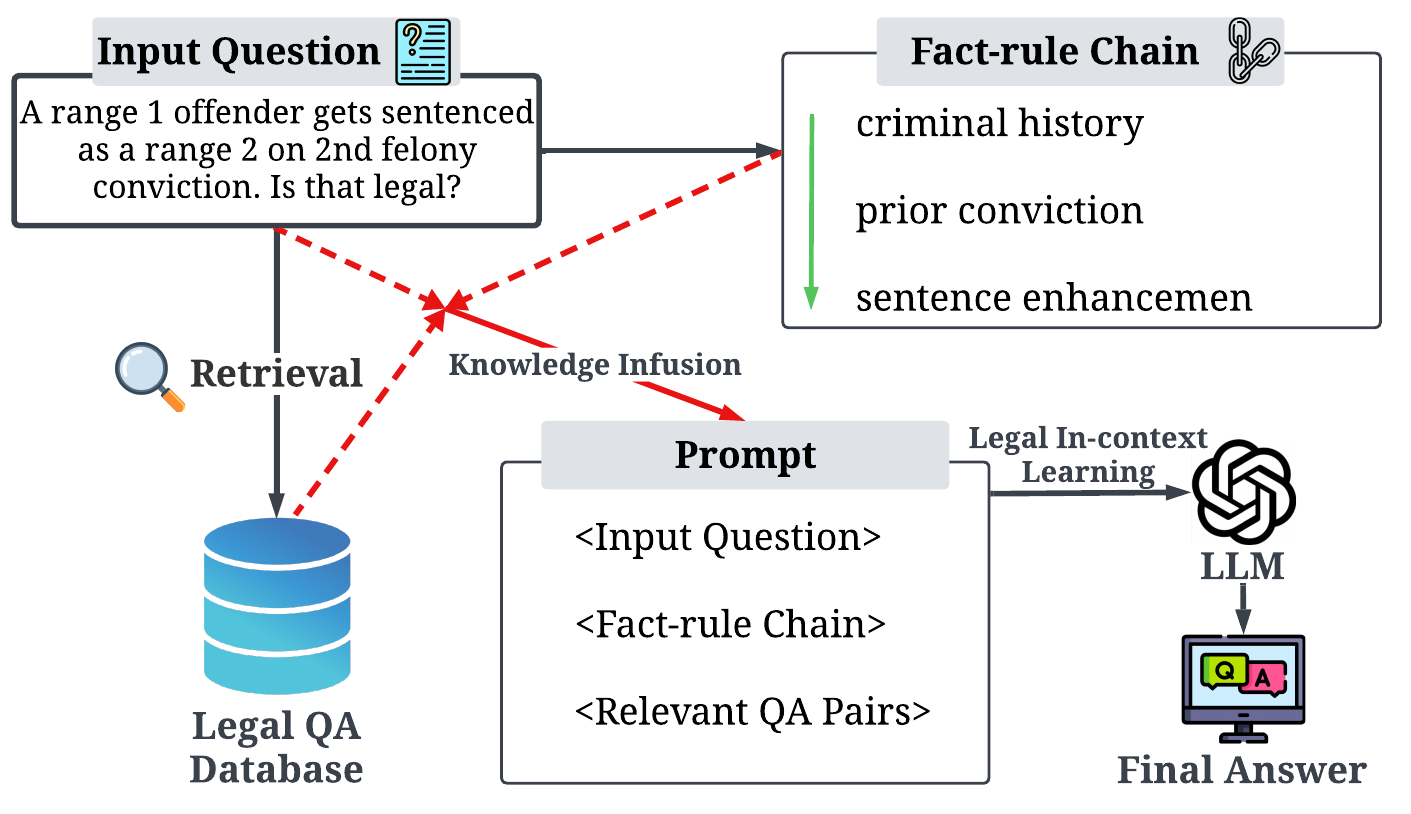}
    \caption{An illustration of our model.}
    \label{fig:fig1}
\end{figure}
The rapid advancement of Large Language Models (LLMs) has provided ordinary individuals with a way to access affordable legal services, substantially expanding their opportunities to obtain legal assistance~\citep{cheong2024not,louis2024interpretable}. However, given the diversity, complexity, and vague expressions of legal questions, responses generated by LLMs often lack logical specificity and may include hallucinations, exposing users to potential legal risks and financial losses~\citep{dahl2024large}. Retrieval-Augmented Generation (RAG)~\citep{lewis2020retrieval,li2024enhancing} has emerged as a promising approach to mitigate hallucinations and improve response accuracy by retrieving relevant legal cases and provisions as contextual knowledge sources for LLMs. Nevertheless, existing RAG methods primarily focus on semantic similarity and frequently overlook the intricate logical structures and reasoning essential for addressing complex legal issues, limiting their effectiveness in real-world legal applications.

As illustrated in Fig.~\ref{fig:fig1}, this paper introduces a novel LLM framework specifically designed to address complex legal QA tasks. Our approach integrates learnable chain-of-thought (CoT) reasoning as logical structures with supervised RAG, enhanced by the Deep Structured Semantic Model (DSSM) \citep{huang2013learning}. By embedding logical reasoning into the retrieval process, the framework ensures that both semantic relevance and logical coherence are maintained. Additionally, in-context learning is leveraged to synthesize high-quality answers, utilizing the retrieved auxiliary information to deliver precise, contextually appropriate responses.

The contributions of this work are fourfold:

\begin{itemize}
  \item We propose a novel LSIM framework, which consists of three components: learnable fact-rule chain, supervised DSSM-powered RAG, and legal in-context learning for precise legal answer generation.
  \item We employ reinforcement learning to estimate the logical structure of users' legal questions, which integrates both logical structures and semantic information, and navigates LLMs to generate responses like legal professional.
  \item We extract fact-rule information in the form of chain of thought from users' legal questions, identifying essential facts and applicable legal rules. This enables the system to understand complex legal issues with precision, facilitating the retrieval of highly relevant case law.
  \item We conducted extensive experiments on a real-world legal QA dataset collected specifically for this study. The results validate the effectiveness and reliability of the proposed framework.
\end{itemize}

\section{Related work}

\subsection{Retrieval-Augmented Generation (RAG)}

RAG can significantly improve the model performance by leveraging additional knowledge and has been widely applied in various tasks, such as question \& answering (Q\&A)~\citep{lewis2020retrieval, mao2020generation}, machine
translation~\citep{gu2018search}, and summarization~\citep{liu2020retrieval, parvez2021retrieval}. With the emergence of LLMs such as LLaMA and ChatGPT, the integration of RAG with LLMs has gained significant popularity and led to significant advancements in multiple tasks~\citep{liu2023reta, kim2023tree, sharma2024retrieval, feng2024retrieval}.

RAG is also widely applied in research within the legal domain, such as legal Q\&A~\citep{cui2023chatlaw, louis2024interpretable, wiratunga2024cbr}, legal judgment prediction~\citep{wu2023precedent}, legal text evaluation~\citep{ryu2023retrieval}, and terminology drafting for legislative documents~\citep{chouhan2024lexdrafter}. 

However, most prior research primarily concentrates on improving the performance of retrieval models from a semantic perspective. While semantic information is undoubtedly important, the significance of logical structure is particularly prominent in dealing with legal questions. Legal reasoning often relies on a well-defined logical flow. To address this challenge, our study emphasizes the integration of both semantic information and logical structure in retrieval processing.

\subsection{Question \& Answering (Q\&A)}
Q\&A is an active research area in NLP that aims to develop systems capable of providing accurate and relevant answers to questions posed in natural language by users based on large knowledge sources~\citep{Rogers2023}. Current Q\&A studies mainly focus on 1) knowledge retrieval which aims to develop effective and efficient methods to retrieve relevant information from large knowledge bases or corpora~\citep{Vladimir2020}, 2) reading comprehension which aims to build models that can comprehend passages to identify answer-relevant information~\citep{Baradaran2022}, 3) multi-hop reasoning, which aims to perform multi-step reasoning by combining information from multiple sources~\citep{wang-etal-2022-new}, and 4) explainable Q\&A which aims to generate human-understandable explanations or rationales to support their answers~\citep{Veronica2020}.

\subsection{AI Applications in Law}
The legal domain has seen increasing interest in applying AI and machine learning techniques to assist with various tasks in Law. One active area of research is using NLP for legal document analysis and information extraction~\citep{Zhong2020}.~\citet{mistica2020} created a schema based on related information that legal professionals seek within judgements and performed classification based on it.~\citet{Sun-xu-2023} proposed a model-agnostic causal learning framework to for legal case matching. There is also work on using AI for legal judgment prediction, as in~\citet{Liu-Zhang-2023} who develop a neural framework to predict judgments from fact descriptions.

Another emerging application is using AI for legal QA, legal reasoning, and argument mining from texts.~\citet{chen2023equals} proposed a well-annotated real-world dataset for legal QA.~\citet{Mumford2023} establihsed a new dataset and explored neural methods to capture patterns of reasoning over legal texts.~\citet{Zhang-Nulty-2023} investigated extracting argumentative components like claims and evidence from legal cases. Some researchers are also exploring constitutionality analysis, with~\citet{Sert2022} proposing an AI system to predict decisions of the Turkish constitutional court. While promising, these AI-based legal methods still face challenges around interpretability, generalization, and capturing the nuanced reasoning required in law.

\section{Methodology}
\begin{figure*}[htbp]
\setlength{\belowcaptionskip}{-0.3cm}
    \centering
    \includegraphics[width=0.95\linewidth]{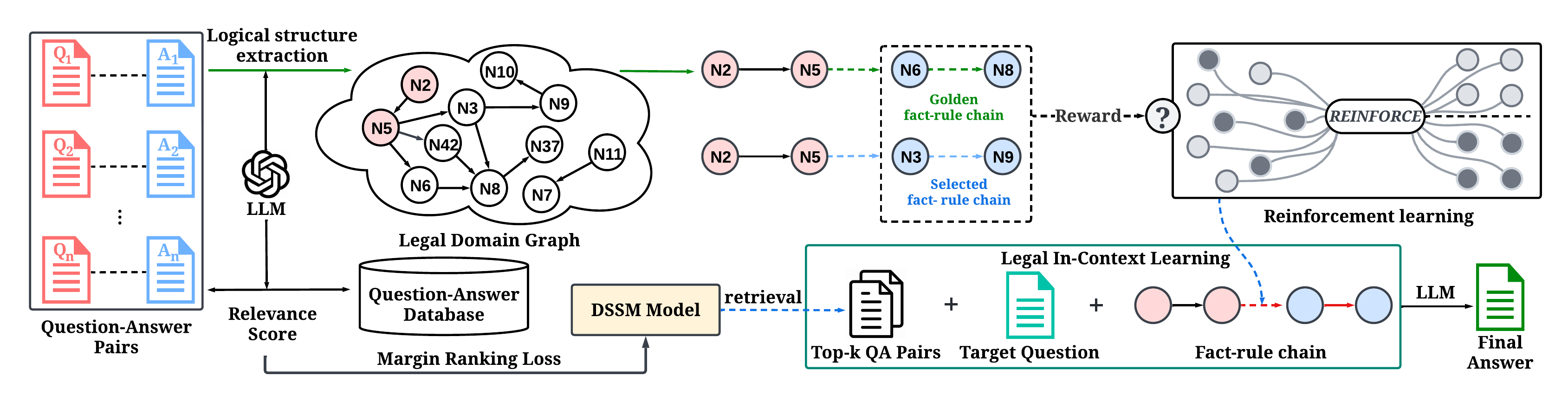}
    \caption{The overall framework of LSIM. The network consists of three modules: learnable fact-rule chain, supervised DSSM-powered RAG, and legal in-context learning.}
    \label{fig:fig2}
\end{figure*}

In this section, we propose our Logical-Semantic Integration Model (LSIM), as depicted in Fig.~\ref{fig:fig2}.

\subsection{Learnable fact-rule chain}
In the legal domain, judging a case requires comprehensive consideration of the facts of the case and relevant legal rules. Each judgment process is a reasoning process that needs to combine the facts of the case with legal rules to reach a final conclusion. The user's question and the lawyer's response can be viewed as a complete legal case. Constructing a fact-rule chain for the case helps clarify the entire logical structure and more clearly identify the core issues for the case. Therefore, in this study, we represent the logical structure as the fact-rule chain of the question and its answer. Each node in the fact-rule chain consists of either a fact or a rule. Fact nodes correspond to the specific circumstances of the case, such as illegal search, arrest, and indictment. Rule nodes correspond to the relevant legal basis applicable to legal case circumstances, such as Fourth Amendment, harmless error doctrine, and Federal Rules of Evidence. The complete fact-rule chain illustrates the comprehensive reasoning process for the legal question/answer.

\subsubsection{Logical structure extraction}

Following~\citet{wu2024knowledge}, a fact-rule graph $\mathcal{G}$ is constructed using the LLM. Assume our training set $T= \{(q_i, a_i)\}_{i=1}^N$ contains $N$ instances, where $q_i$ is the $i$-th question, and $a_i$ is the real lawyer's answer to $ q_i$. For each question-answer pair $(q_i, a_i) $, the LLM, guided by a tailored prompt, is employed to extract the most relevant fact-rule chain from graph $\mathcal{G}$. The prompt is provided in Appendix~\ref{sec:appendix01}\footnote{In cases where an exact fact/rule node cannot be found, the most similar node is selected.}. Then, the fact-rule chain $C_{q_i}$ for question $q_i$ is obtained, and  $C_{q_i} = \{c_{q_i,1}, c_{q_i,2}, ..., c_{q_i,t}\}$ where $c_{q_i,t}$ is the $t$-th fact-rule chain node of the question $q_i$. Similarly, the fact-rule chain $C_{a_i}$ for answer $a_i$ is obtained, and $C_{a_i} = \{c_{a_i,1}, c_{a_i,2}, ..., c_{a_i,t}\}$ where $c_{a_i,t}$ is the $t$-th fact-rule chain node of the answer $a_i$\footnote{Based on our analysis and statistics, we have identified that a maximum of four key elements are sufficient to cover the main content of a legal question/answer. Therefore, in this paper, the value of $t$ ranges from 1 to 4.}. Consequently, the fact-rule chain for all questions $C_Q$ and the fact-rule chain for all answers $C_A$ in the training set are obtained, where $C_Q= \{C_{q_i}\}_{i=1}^N$ and $C_A= \{C_{a_i}\}_{i=1}^N$\footnote{$C_Q$ is a set composed of fact-rule chains for $N$ legal questions, and $C_A$ is a set composed of fact-rule chains for $N$ legal answers.}.

\subsubsection{Learning to predict fact-rule chain}\label{CoT_Prediction}
The prediction of fact-rule chain is inherently a cumulative and continuous process. Therefore, we consider the fact-rule chain prediction task as a sequential decision-making process, and a reinforcement learning-based approach is empolyed. Specifically, reinforcement learning is employed to predict the fact-rule chain of the lawyer's response based on the information detected from the user's question. This prediction serves to supplement the information in the user’s question. Given the fact-rule chain $C_{q_i}$ for the legal question $q_i$, we first encode $C_{q_i}$ using BERT~\citep{kenton2019bert} to obtain its embedding representation:

\begin{equation} 
h_{C_{q_i}}= Encode(C_{q_i}).
\end{equation}

Then, we utilize a policy network $\pi_{\theta}(n_t|s_t)$ to predict the fact-rule chain for answer $a_i$ step by step, where $s_t$ represents the current state at time step $t$, and $n_t$ denotes the action (next fact-rule chain node) predicted by the policy network. The initial fact-rule chain $C^{t=0}_{q_i}$ is set to $C{q_i}$, and the initial state $s_0$ is set to $h_{C_{q_i}}$. At step $t$, the policy network selects an action $n_t$ based on the current state $s_t$. Then, the selected node $n_t$ is appended to the current fact-rule chain:
\begin{equation} 
C^{t+1}_{q_i} = [C^t_{q_i},n_t],\quad n_t \sim \pi_{\theta}(n_t|s_t)
\end{equation}
Subsequently, the state embedding is updated using the new fact-rule chain:
\begin{equation} 
s_{t+1} =Encode(C^{t+1}_{q_i}).
\end{equation}

This process is repeated until a maximum number of steps is reached or no valid next node can be selected. The policy network is implemented as a multi-layer perceptron (MLP). The REINFORCE algorithm~\citep{williams1992simple} is employed to train the policy network, which is a classic policy gradient method in reinforcement learning. The REINFORCE algorithm focuses strongly on maximizing long-term rewards. This enables the model to consider more far-reaching goals and impacts, rather than merely immediate predictive performance. In complex tasks such as legal reasoning, each thought step is built upon the foundation of previous logical reasoning. The REINFORCE algorithm can effectively simulate this process, learning the logical relationships between nodes in the fact-rule chain, thereby ensuring that the generated fact-rule chain is logically coherent and accumulative. The training objective is to maximize the expected cumulative reward:
\begin{equation}
J(\theta) = \mathbb{E}{\pi{_\theta}}[\sum_{t=0}^{T} r_t],
\end{equation}
where $r_t$ is the reward at step $t$, and $T$ is the maximum number of steps. The reward $r_t$ is defined as follows:
\begin{equation}
r_t = \begin{cases}
1, & \text{if } n_t \in C_{a_i} \\
0, & \text{otherwise}
\end{cases}
\end{equation}
where $C_{a_i}$ is the ground-truth fact-rule chain for answer $a_i$.

During inference, the trained policy network is employed to predict the fact-rule chain for a given legal question. Assume the inference step is $z$, the predicted fact-rule chain is $C^z_{q_i}$ for question $q_i$, and $C^z_{q_i}$ is the predicted logical structure. Appendix~\ref{sec:appendix1} provides examples illustrating the fact-rule chains.

\subsection{Supervised DSSM-powered RAG}
Deep Structured Semantic Model (DSSM)~\citep{huang2013learning} is utilized to retrieve the most relevant questions from the database that are relevant to the user's question $q_i$ in terms of legal knowledge. These retrieved questions, along with corresponding responses from lawyers, are provided to the LLM as context, assisting it in generating more accurate responses to the current user's question.

Let $D$ be the database of candidate questions, and $D= \{(q^D_j, a^D_j)\}_{j=1}^M$ contains $M$ instances, where $q^D_j$ is the $j$-th candidate question in $D$, and $a^D_j$ is the real lawyer's answer to $q^D_j$.

Given a legal question $q_i \in T$,  its logical structure $C^z_{q_i}$ can be obtained by the method described in Section~\ref{CoT_Prediction}. Similarly, for each candidate question $q^D_j \in D$,  its logical structure $C^z_{q^D_j}$ can also be obtained. Then we encode each of them independently using the same encoder:

\begin{align}
h_{q_i}= Encode(q_i), \notag \\
h_{C_{q_i}}= Encode(C^z_{q_i}), \notag \\
h_{q^D_j}= Encode(q^D_j), \notag \\
h_{C_{q^D_j}}= Encode(C^z_{q^D_j}).
\end{align}

Subsequently, $h_{C_{q_i}}$, which represents the logical structure features, and $h_{q_i}$, which represents the semantic features, are concatenated together to form the features for the current question $q_i$:
\begin{equation}
e_{q_i} = [h_{C_{q_i}}, h_{q_i}].
\end{equation}
Similarly, the features for candidate question $q^D_j $ can be obtained:
\begin{equation}
e_{q^D_j} = [h_{C_{q^D_j}}, h_{q^D_j}].
\end{equation}

The DSSM model is composed of a multi-layer perceptron (MLP) and computes a relevance score $p_{ij}$ between $q_i$ and candidate question $q^D_j$:
\begin{align}
x_1 &= \tanh(W_1 [e_{q_i}, e_{q^D_j}] + b_1) \notag\\
x_2 &= \tanh(W_2 x_1 + b_2) \notag\\
x_3 &= \tanh(W_3 x_2 + b_3)\notag \\
p_{ij} &= W_4 x_3 + b_4,
\end{align}
where $W_1, W_2, W_3$, and $W_4$ are weights, and $b_1, b_2, b_3$ , and $b_4$ are bias.

The margin ranking loss is employed, which encourages the model to assign higher scores to more relevant cases. For each question $q_i$, we select the candidate question in the database with the highest annotated relevance score as the positive example $c_i^+$, and the candidate question with the lowest score as the negative example $c_i^-$. The annotated relevance scores are generated by the LLM. Specifically, the LLM evaluates the relevance between the current query and each candidate question. These relevance scores are assigned on a scale from 0 to 5, where a score of 0 indicates minimal relevance and a score of 5 denotes the highest level of similarity.The prompt is provided in Appendix~\ref{sec:appendix015} The loss function is defined as:
\begin{equation}
\mathcal{L}(q_i, c_i^+, c_i^-) = \max(0, \alpha - p(q_i, c_i^+) + p(q_i, c_i^-)),
\end{equation}
where $\alpha$ is a hyperparameter.

During inference, for each question $q_i$, we compute the relevance scores between $q_i$ and all candidate questions in the database $D$ using the trained DSSM model. The top-K candidate questions with the highest scores are the final retrieval results.

\subsection{Legal in-context learning}
After retrieving the top-K most relevant questions ${q^D_{j_1}, q^D_{j_2}, ..., q^D_{j_K}}$ from the database $D$ for the current question $q_i$, we concatenate them with their corresponding answers ${a^D_{j_1}, a^D_{j_2}, ..., a^D_{j_K}}$ to form the context for in-context learning:
\begin{equation}
context_i = [(q^D_{j_1}, a^D_{j_1}), (q^D_{j_2}, a^D_{j_2}), ..., (q^D_{j_K}, a^D_{j_K})].
\end{equation}
This context provides the LLM with relevant examples of how relevant legal questions have been answered by real lawyers in the past. Following~\citet{wu2023precedent} and considering the length limit of the prompt, K is set to 3 in our experiments. 

Then, the current question $q_i$, the logical structure $C_{q_i}$, and $context_i$ are provided to the LLM to generate an answer. The prompt is provided in Appendix~\ref{sec:appendix02}:
\begin{equation}
a_i^\prime  = LLM(q_i,C_{q_i}, context_i).
\end{equation}

\section{Experiments}

\subsection{Datasets}

We use real-world legal question and answer (Q\&A) data collected from JUSTIA\footnote{https://www.justia.com/}. The dataset comprises 16,190 legal questions posed by users in the field of criminal law, with each question receiving responses from an average of 1.26 lawyers. The average length of the questions is 67 words, and the responses average 40 words. In the experiment, the total dataset comprising 16,190 samples is divided using an 8:2 ratio: 80\% of the data (12,952 samples) is used as the database for retrieval, and 20\% (3,283 samples) is used for training and testing. Subsequently, the 3,238 samples are further divided into training and testing sets using the same 8:2 ratio. The specific information is presented in Table~\ref{data_details}. Appendix~\ref{sec:appendix2} presents some samples of the collected data\footnote{The code and data are available at~\url{https://github.com/RujingYao/LSIM}.}.

\begin{table}[ht]
\centering
\begin{tabular}{|c|c|}
\hline
\textbf{Type} & \textbf{Value} \\ \hline
\multicolumn{2}{|c|}{\textbf{Data Characteristics}} \\ \hline
Total number of questions & 16,190 \\ \hline
Total number of responses & 20,400 \\ \hline
Mean length of questions (words) & 67 \\ \hline
Mean length of responses (words) & 40 \\ \hline
\multicolumn{2}{|c|}{\textbf{Data Split}} \\ \hline
Database & 12,952 \\ \hline
Training & 2,590 \\ \hline
Testing & 648 \\ \hline
\end{tabular}
\caption{\label{data_details}Statistics of data we collected.}
\end{table}

\subsection{Baselines and evaluation metrics}
\textbf{Baselines.} We implement the following baselines for comparison:
\textbf{BM25}~\citep{robertson1994some}, a classic bag-of-words information retrieval model, is used to retrieve the question from the database that most closely matches the user's query. \textbf{Bert}~\citep{kenton2019bert}, \textbf{Roberta}~\citep{liu2019roberta}, \textbf{Sentence T5}~\citep{ni2022sentence}, \textbf{INSTRUCTOR}~\citep{su2023one}, \textbf{GTR}~\citep{ni2022large}, \textbf{BGE-m3}~\citep{chen2024bge}, \textbf{text-embedding-ada-002}, ~\textbf{text-embedding-3-small}, and ~\textbf{text-embedding-3-large}, are employed to generate embeddings for the user's query and questions in the database. Similarity calculations are then used to determine the closest match. Following~\citet{louis2024interpretable, wu2023precedent, wu2024knowledge},~\textbf{LLaMA-2-13B},~\textbf{LLaMA-3-8B}\footnote{https://llama.meta.com/}, and~\textbf{GPT-4o}, serve as the LLM baselines in our study. They can generate responses to the posed questions.

\begin{table*}[h]
\footnotesize
\centering
\resizebox{\linewidth}{!}{
\begin{tabular}{cccccc}
\toprule
Method                    & METEOR            &  ROUGE-1           & ROUGE-2           & ROUGE-L    & BERTScore     \\
\midrule
LLaMA-2-13B w/o RAG        & 17.09      & 13.13   & 1.91    & 12.09   & 81.24 \\
BM25                       & 17.75      & 13.98   & 2.22    & 12.81   & 81.10  \\
Bert-Base                  & 17.96      & 13.94   & 2.18    & 12.69   & 81.33  \\
Roberta                    & 17.87      & 13.85   & 2.21    & 12.71   & 81.27  \\
SentenceT5-Base            & 17.86      & 13.99   & 2.28    & 12.85   & 81.74  \\
INSTRUCTOR-Base            & 17.92      & 14.12   & 2.26    & 12.82   & 81.81  \\
GTR-Base                   & 17.76      & 13.95   & 2.22    & 12.81   & 81.58  \\
BGE-m3                     & 17.85      & 13.75   & 2.14    & 12.59   & 81.35  \\
text-embedding-ada-002     & 17.76      & 13.89   & 2.22    & 12.70   & 81.45  \\
text-embedding-3-small     & 17.89      & 14.02   & 2.16    & 12.76   & 81.53  \\ 
text-embedding-3-large     & 18.03      & 14.13    & 2.23    & 12.93   & 81.62  \\  \midrule
LSIM             &{\bf 20.55}   &{\bf 16.10}   &{\bf 2.58}  &{\bf 14.52} &{\bf 83.12}\\ \bottomrule
\end{tabular}}
\caption{Performance on legal response generation using LLaMA-2-13B (\%).}
\label{tab:results_llama2}
\end{table*}

\begin{table*}[htbp]
\footnotesize
\centering
\resizebox{\linewidth}{!}{
\begin{tabular}{cccccc}
\toprule
Method                    & METEOR           &  ROUGE-1           & ROUGE-2           & ROUGE-L     & BERTScore     \\
\midrule
LLaMA-3-8B w/o RAG               & 17.13     & 11.56   & 1.69    & 10.62   & 81.91  \\
BM25                     & 17.84     & 13.34   & 2.09    & 12.10   & 82.47  \\
Bert-Base                & 17.68     & 13.13   & 1.99    & 11.92   & 82.53  \\
Roberta                  & 18.03     & 13.38   & 2.14    & 12.15   & 82.56  \\
SentenceT5-Base          & 18.40     & 13.58   & 2.16    & 12.40   & 82.50  \\
INSTRUCTOR-Base          & 18.39     & 13.61   & 2.28    & 12.29   & 82.49  \\
GTR-Base                 & 18.10     & 13.54   & 2.15    & 12.32   & 82.48  \\
BGE-m3                   & 18.08     & 13.46   & 2.08    & 12.12   & 82.52  \\
text-embedding-ada-002   & 17.87     & 13.24   & 2.04    & 12.01   & 82.48  \\
text-embedding-3-small   & 18.32     & 13.70   & 2.22    & 12.44   & 82.49  \\ 
text-embedding-3-large   & 18.62     & 13.82   & 2.24    & 12.53   & 82.52  \\ \midrule
LSIM              &{\bf 21.00}   &{\bf 16.30}   &{\bf 2.63}  &{\bf 14.74} &{\bf 83.23}\\ \bottomrule
\end{tabular}}
\caption{Performance on legal response generation using LLaMA-3-8B (\%).}
\label{tab:results_llama3}
\end{table*}

\begin{table*}[htbp]
\footnotesize
\centering
\resizebox{\linewidth}{!}{
\begin{tabular}{cccccc}
\toprule
Method                    & METEOR           &  ROUGE-1           & ROUGE-2           & ROUGE-L     & BERTScore     \\
\midrule
GPT-4o w/o RAG           & 17.92     & 12.36   & 2.20    & 11.46   & 81.60  \\
BM25                     & 18.41     & 12.88   & 2.22    & 11.84   & 82.01  \\
Bert-Base                & 18.53     & 12.90   & 2.24    & 11.91   & 82.11  \\
Roberta                  & 18.50     & 13.01   & 2.24    & 11.94   & 82.16  \\
SentenceT5-Base          & 18.57     & 12.97   & 2.23    & 11.90   & 82.05  \\
INSTRUCTOR-Base          & 18.55     & 12.92   & 2.30    & 11.92   & 82.04  \\
GTR-Base                 & 18.40     & 12.84   & 2.25    & 11.80   & 82.00  \\
BGE-m3                   & 18.51     & 12.96   & 2.25    & 11.88   & 82.05  \\
text-embedding-ada-002   & 18.35     & 12.65   & 2.23    & 11.73   & 81.98  \\
text-embedding-3-small   & 18.47     & 12.88   & 2.21    & 11.86   & 81.97  \\ 
text-embedding-3-large   & 18.69     & 13.02   & 2.31    & 12.01   & 82.04  \\ \midrule
LSIM              &{\bf 21.54}   &{\bf 14.09}   &{\bf 2.31}  &{\bf 12.94} &{\bf 82.68}\\ \bottomrule
\end{tabular}}
\caption{Performance on legal response generation using GPT-4o (\%).}
\label{tab:results_gpt4o}
\end{table*}

\begin{table}[htbp]
\centering
\resizebox{\linewidth}{!}{
\begin{tabular}{lccc}
\toprule
Method       & Acc. & Spec. & Adopt. \\
\midrule
LLaMA-3-8B & 4.08  & 4.33 & 4.25 \\
text-embedding-3-large     & 4.35  & 4.35 & 4.41 \\
LSIM          & 4.65  & 4.47 & 4.65  \\
\bottomrule
\end{tabular}}
\caption{Results of human evaluation.}\vspace{-0.2in}
\label{tab:human_evaluation}
\end{table}

\begin{table*}[htbp]
\centering
\resizebox{\textwidth}{!}{
\small
\begin{tabular}{|p{0.5\linewidth}|p{0.5\linewidth}|}
\hline
\multicolumn{2}{|c|}{\textbf{User’s Question}} \\
\hline
\multicolumn{2}{|p{\linewidth}|}{Someone I know has been accused of touching a child. She is 18 now and is making these allegations. How can he clear himself. A friend of mine has been accused of inappropriately touching a young child when she was six, that‘s what she is saying. She is 17 or 18 now. My friend is very upset and is wrongfully accused. Can he take her to court to get his name cleared? If so, what steps should he go about this?} \\
\hline
\multicolumn{2}{|c|}{\textbf{The most relevant questions retrieved by LSIM}} \\
\hline
\multicolumn{2}{|p{\linewidth}|}{A false allegation of inappropriate touching was made and a polygraph is being requested is there a way to dismiss this? My teen step-daughter has a history of bad behavior and being unruly. During a recent counseling session, she accused me of touching her while giving her a hug last June. CPS is involved and I'm now being asked to take a polygraph. Since the incident, she has been sent to stay with her grandmother after sneaking out and breaking a neighbors window.} \\
\hline
\multicolumn{2}{|c|}{\textbf{Lawyer's Answer}} \\
\hline
\multicolumn{2}{|p{\linewidth}|}{I recommend you \textcolor{red}{keep your mouth shut} and do not post anything else online.  \textcolor{red}{Hire a competent attorney} today to counsel you on your possible criminal charges and how to conduct yourself during this DCS and/or LEO investigation. Again do not talk to anyone about this and have no contact with the girl.} \\
\hline
\end{tabular}}
\caption{The most relevant questions retrieved by LSIM.}
\label{tab:relevantcase}
\end{table*}

\begin{table*}[htbp]
\centering
\begin{tabular}{cccccc}
\toprule
Method                    & METEOR          &  ROUGE-1           & ROUGE-2           & ROUGE-L     & BERTScore      \\
\midrule
LSIM     &{\bf 21.00}   &{\bf 16.30}   &{\bf 2.63}  &{\bf 14.74} &{\bf 83.23}\\ 
LSIM w/o LS     & 19.10      & 14.18   & 2.22   & 12.88     & 82.45  \\ 
LSIM w/o SI     & 18.77     & 13.83   & 2.17   & 12.62     & 82.48 \\ \bottomrule
\end{tabular}
\caption{Ablation study when using LLaMA-3-8B as the LLM architecture (\%).}
\label{tab:ablation}
\end{table*}

\textbf{Evaluation metrics.} 
To evaluate our model, both automatic and human evaluations are used. For automatic evaluation, the commonly used text generation metrics, \textbf{ROUGE} (ROUGE-1, ROUGE-2, and ROUGE-L)~\citep{lin2004rouge}, \textbf{METEOR}~\citep{banerjee2005meteor}, and \textbf{BERTScore}~\citep{zhang2019bertscore} are employed. Human evaluation focuses on three aspects: 1)  \textbf{Accuracy}: the aspect evaluates whether the generated answers are correct and free from factual errors.  2)  \textbf{Specificity}: this aspect measures whether the responses are directly related to the specific issues raised in the question, providing clear and targeted answers rather than generalized responses. 3)  \textbf{Adoptability}: this aspect assesses whether the responses generated by the model are practically useful and can be directly provided to users. 

Three legal professionals were invited to evaluate the answers generated by LLaMA-3-8B, text-embedding-3-large, and our proposed Model LSIM. Each dimension is rated on a scale of 1-5, with 5 being the highest score. To ensure a fair and unbiased evaluation, each of the three professionals was presented with outputs from all tested LLMs in a randomized order, with no indication of which outputs corresponded to which models. The three professionals independently scored all the results, and the average score was calculated.

\begin{figure*}[ht]
\setlength{\belowcaptionskip}{-0.3cm}
    \centering
    \includegraphics[width=0.9\linewidth]{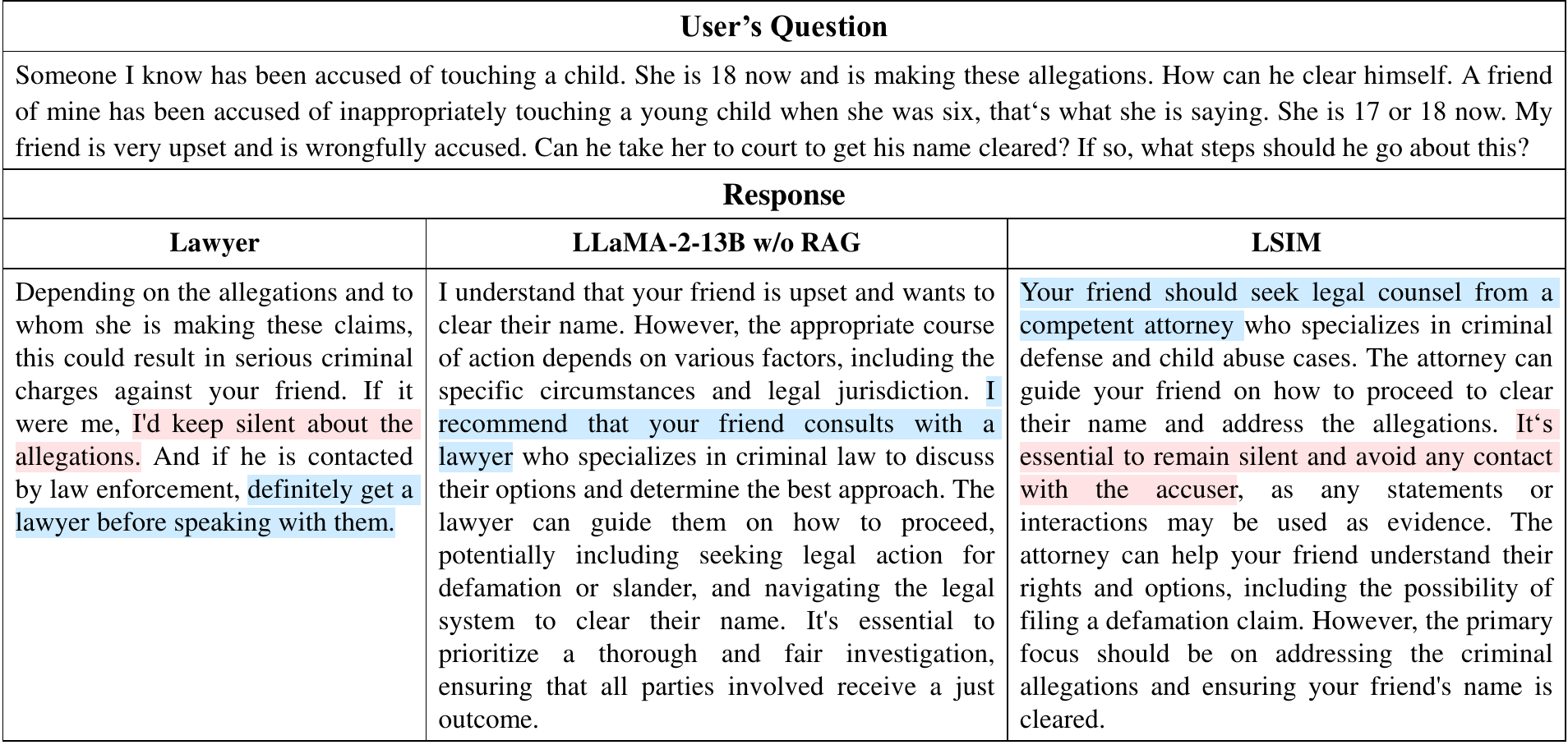}
    \caption{Illustrative example 1. The sentences highlighted in red indicate remaining silent, while those highlighted in blue indicate seeking professional legal counsel.}
    \label{fig:fig3}
\end{figure*}

\begin{figure*}[ht]
\setlength{\belowcaptionskip}{-0.3cm}
    \centering
    \includegraphics[width=0.80\linewidth]{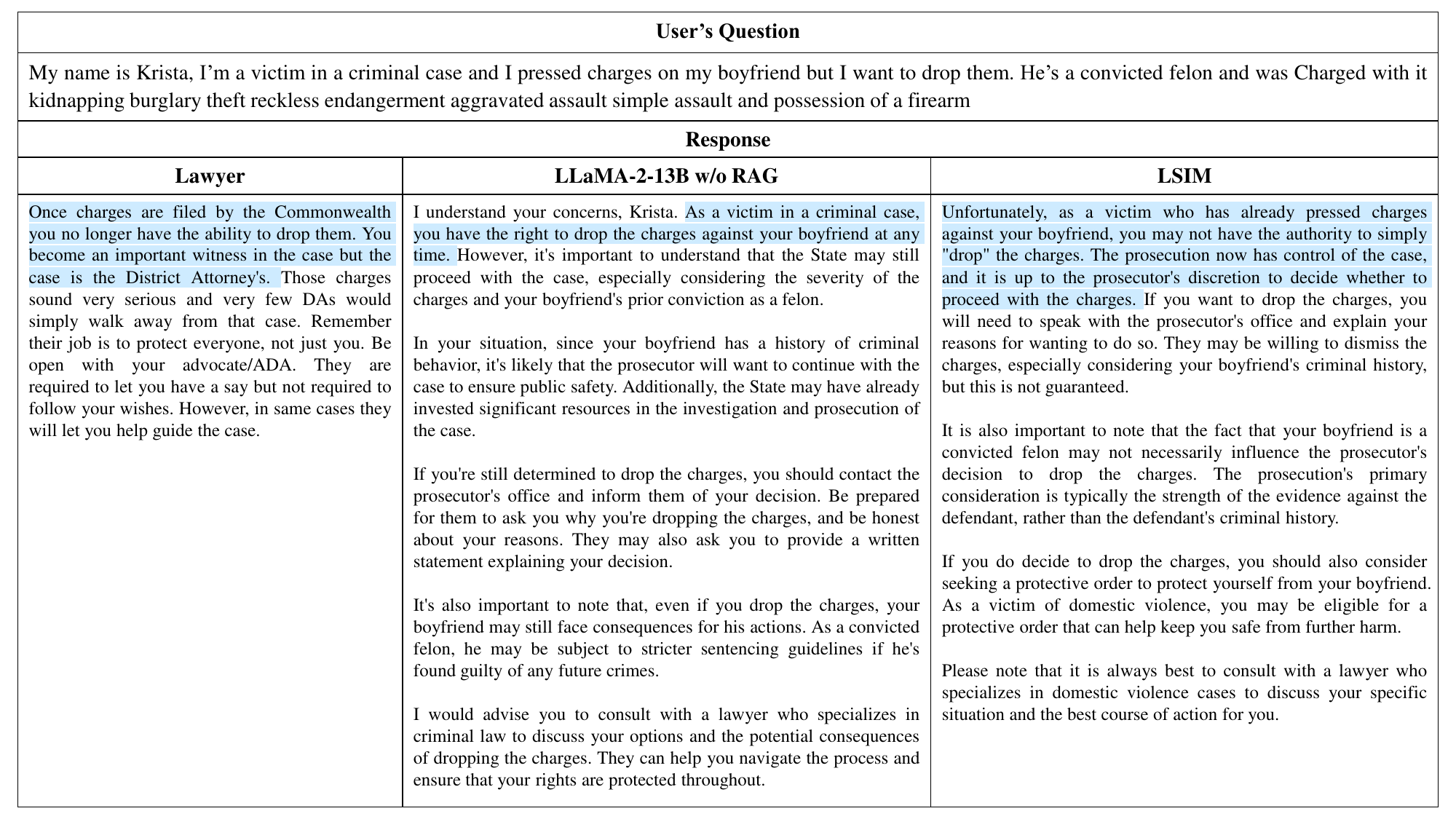}
    \caption{Illustrative example 2. The sentences highlighted in blue indicate responses from different models regarding the user's right to drop charges.}
    \label{fig:fig4}
\end{figure*}

\begin{figure*}[ht]
\setlength{\belowcaptionskip}{-0.3cm}
    \centering
    \includegraphics[width=0.80\linewidth]{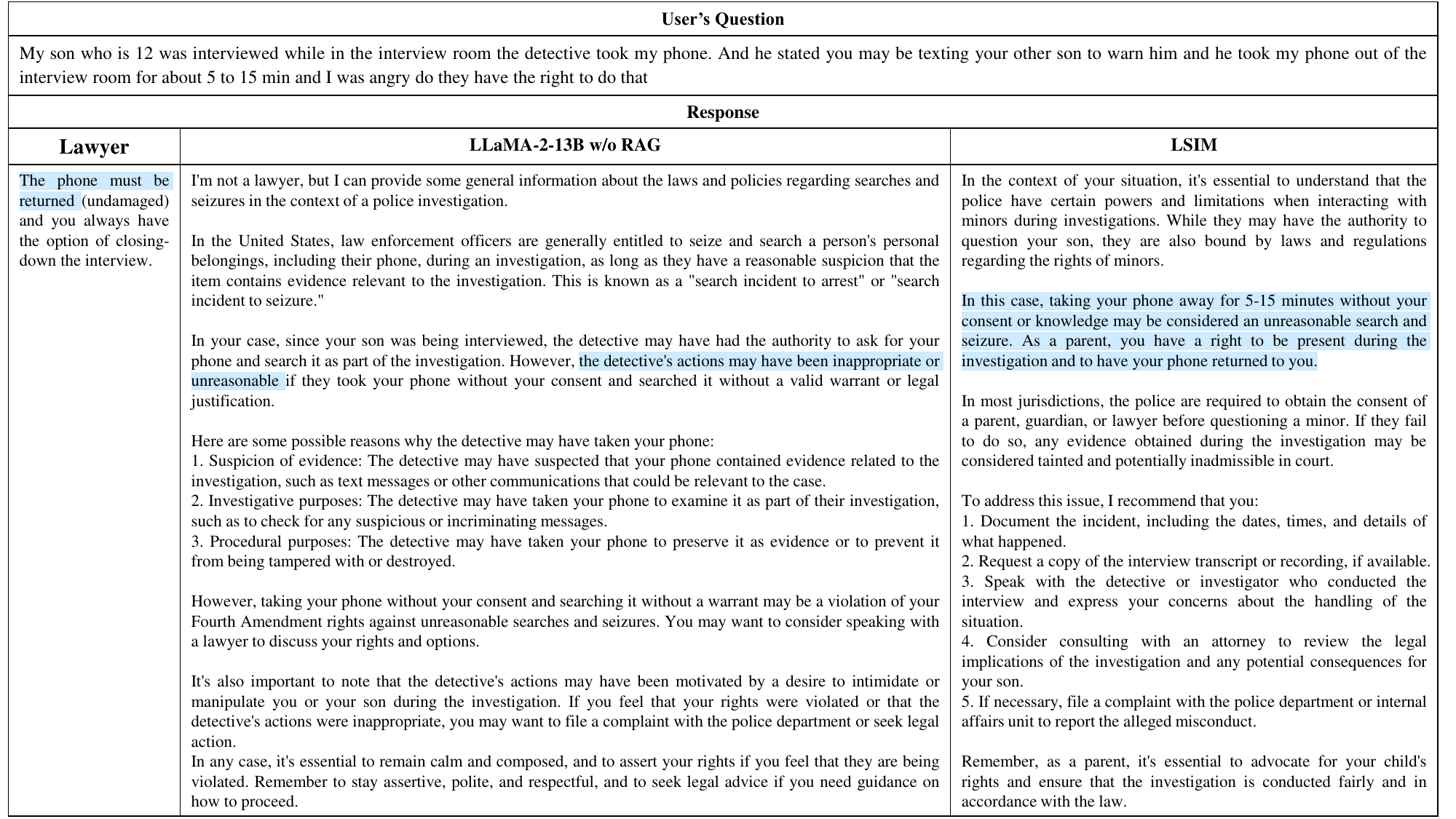}
    \caption{Illustrative example 3. The sentences highlighted in blue indicate responses from different models regarding the user's right to property recovery.}
    \label{fig:fig5}
\end{figure*}
\subsection{Experiment Settings}
For GPT-4o, the sampling parameters are set with a temperature of 0.8. For LLaMA-2-13B and LLaMA-3-8B, the sampling parameters are set with a temperature of 0.8 and a top-p value of 0.9. The maximum token limit per generation is set at 4096. For the LSIM method, the word embeddings are initialized using BERT. Adam is used as the optimizer. The learning rates for the policy network and the DSSM are both set to 1e-4. The number of epochs for the policy network and the DSSM are set to 30 and 50, respectively. All the results are the average values of three repeated runs.

\subsection{Experiment Results}

Tables~\ref{tab:results_llama2},~\ref{tab:results_llama3}, and~\ref{tab:results_gpt4o} present the experimental results obtained when LLaMA-2-13B , LLaMA-3-8B, and GPT-4o are used as base architectures, respectively. When LLaMA-2-13B is employed as the LLM baseline, our proposed LSIM algorithm also achieves the best performance across all metrics. Compared with the LLaMA-2-13B w/o RAG, our proposed LSIM model achieves improvements of 3.46\% on METEOR, 2.97\% on ROUGE-1, 0.67\% on ROUGE-2, 2.43\% on ROUGE-L, and 1.88\% on BERTScore. When LLaMA-3-8B is employed as the LLM baseline, our proposed LSIM algorithm achieves the best performance across all metrics. Compared with the LLaMA-3-8B w/o RAG, our proposed LSIM model achieves improvements of 3.87\% on METEOR, 4.74\% on ROUGE-1, 0.94\% on ROUGE-2, 4.12\% on ROUGE-L, and 1.32\% on BERTScore. Compared to the best performing baseline model text-embedding-3-large, LSIM model achieves improvements of 2.38\%, 2.48\%, 0.39\%, 2.21\%, and 0.71\% on METEOR, ROUGE-1, ROUGE-2, ROUGE-L, and BERTScore. When GPT-4o is employed, compared to the best performing baseline model text-embedding-3-large, LSIM model achieves improvements of 2.85\%, 0.93\%, and 0.64\% on METEOR, ROUGE-L, and BERTScore, respectively. All results demonstrate that our LSIM algorithm achieves the best performance.

\vspace{-0.04in}
The results of the human evaluation are shown in Table \ref{tab:human_evaluation}. Our LSIM model achieves the best performance in terms of accuracy, specificity, and adoptability. These results highlight the effectiveness of our proposed LSIM model.

\subsection{Case Study}

Figures~\ref{fig:fig3},~\ref{fig:fig4} and~\ref{fig:fig5} present the comparison of LLaMA-2-13B directly answering the legal questions (LLaMA-2-13B w/o RAG) and utilizing our LSIM framework to respond to the legal questions. 

For the given question in Figure~\ref{fig:fig3}, there are two main points in a real lawyer's response: keep silent and get a lawyer. However, the response generated by LLaMA-2-13B w/o RAG is comparatively generic, merely mentioning seeking legal counsel. By leveraging our LSIM model, we retrieve relevant questions that are related to the given query. 
Tabel~\ref{tab:relevantcase} presents the most relevant question retrieved by LSIM. The answer to this question advises the user to "keep your mouth shut" and "hire a competent attorney".

By incorporating insights from the retrieved relevant questions, the LSIM model generates a response that covers two crucial aspects: remain silent and seek legal counsel. These key points align closely with the advice given by the real lawyer. The responses generated by our LSIM framework exhibit a higher degree of professionalism and more closely mirror the advice typically provided by a lawyer.

In Figure~\ref{fig:fig4}, the lawyer stated that the user could no longer drop the charges, whereas LLaMA-2-13B w/o RAG responded that the user has the right to drop the charges at any time. The response from our proposed LSIM model aligns with the lawyer's perspective, indicating that the user may not have the authority to drop the charges. 

In Figure~\ref{fig:fig5}, the lawyer responded with, "The phone must be returned," whereas LLaMA-2-13B w/o RAG only noted that "the detective's actions may have been inappropriate or unreasonable." The response from our proposed LSIM model aligns with the key points in the lawyer's reply, explicitly stating that the user has the right to have their phone returned. These examples demonstrate the effectiveness of the proposed LSIM model.

\subsection{Ablation Study}

An ablation study is also conducted on the LSIM framework when using LLaMA-3-8B as base LLM. The results are presented in Tabel~\ref{tab:ablation}. LSIM w/o LS refers to the LSIM model without the Logical Structure module. LSIM w/o SI refers to the LSIM model without the Semantic Information module. The ablation study demonstrates that both the logical structure (LS) and semantic information (SI) modules contribute positively to the overall performance of the LSIM model. The best results are achieved by the full LSIM model, which combines the effects of both the LS and SI components.

\section{Conclusion}
This paper addresses the inherent limitations of LLMs in generating professional legal responses. We propose a novel framework, LSIM, designed to enhance the legal LLM reasoning by integrating the learnable logical structure and semantic information of legal questions. The LSIM framework is composed of three key components: (1) Reinforcement learning predicts the fact-rule chain of thought for a given legal query, guiding the reasoning process; (2) a supervised RAG retrieves the most relevant questions by integrating logical and semantic information; and (3) the fact-rule chain, relevant retrieved questions, and their corresponding answers are provided as auxiliary reference information to the LLM, enabling it to generate precise, contextually relevant responses.

We validate the effectiveness of LSIM through experiments on a real-world legal QA dataset. Results from both automated metrics and manual evaluations demonstrate the superior performance of the proposed framework in delivering accurate and expert-level legal answers.

In the future, we plan to extend the LSIM framework to other specialized domains, such as healthcare and finance. Additionally, we aim to incorporate multi-turn interactions with users, leveraging their real-time feedback to further refine the model's performance and adaptability.

\section*{Limitations}
\label{sec:limitations}
The effectiveness of the RAG-based model heavily depends on the availability of databases. Consequently, our model's performance may degrade due to the lack of sufficient relevant legal cases to retrieve, which hinders the model's adaptability and utility in regions where legal cases are scarce.

As another limitation, our study is limited to single-turn interactions with LLMs. Expanding to multi-turn interactions could enhance the model's ability to develop a more nuanced and comprehensive understanding of user queries. However, achieving this will require a redesign of the proposed model to effectively manage contextual continuity and iterative feedback across multiple exchanges.

\bibliography{custom}

\appendix
\section{Prompt details}
\subsection{Relevant fact-rule path extraction}\label{sec:appendix01}
"Please select 1 to 4 nodes from the provided graph that are most relevant to the legal question/answer. Ensure that the selected nodes are interconnected."
\subsection{Similarity Score}\label{sec:appendix015}
"Please score the similarity of Question2 to Question1, focusing specifically on the events described in each legal question. Rate the similarity of Question2 to Question1 on a scale from 0 to 5, where 0 indicates that Question2 is completely different from Question1, and 5 indicates that Question2 is exactly the same as Question1."

\subsection{Generation of the final response}\label{sec:appendix02}
"Your task is to provide legal advice on the user's question. I will provide you with the logical structure of the user's question, along with similar questions previously asked by other users and the responses given by real lawyers. Please use this information to generate a response to the user's question."

\section{Illustrative examples of the fact-rule chains}

\label{sec:appendix1}
\begin{figure*}[h]
\setlength{\belowcaptionskip}{-0.3cm}
    \centering
    \includegraphics[width=\linewidth]{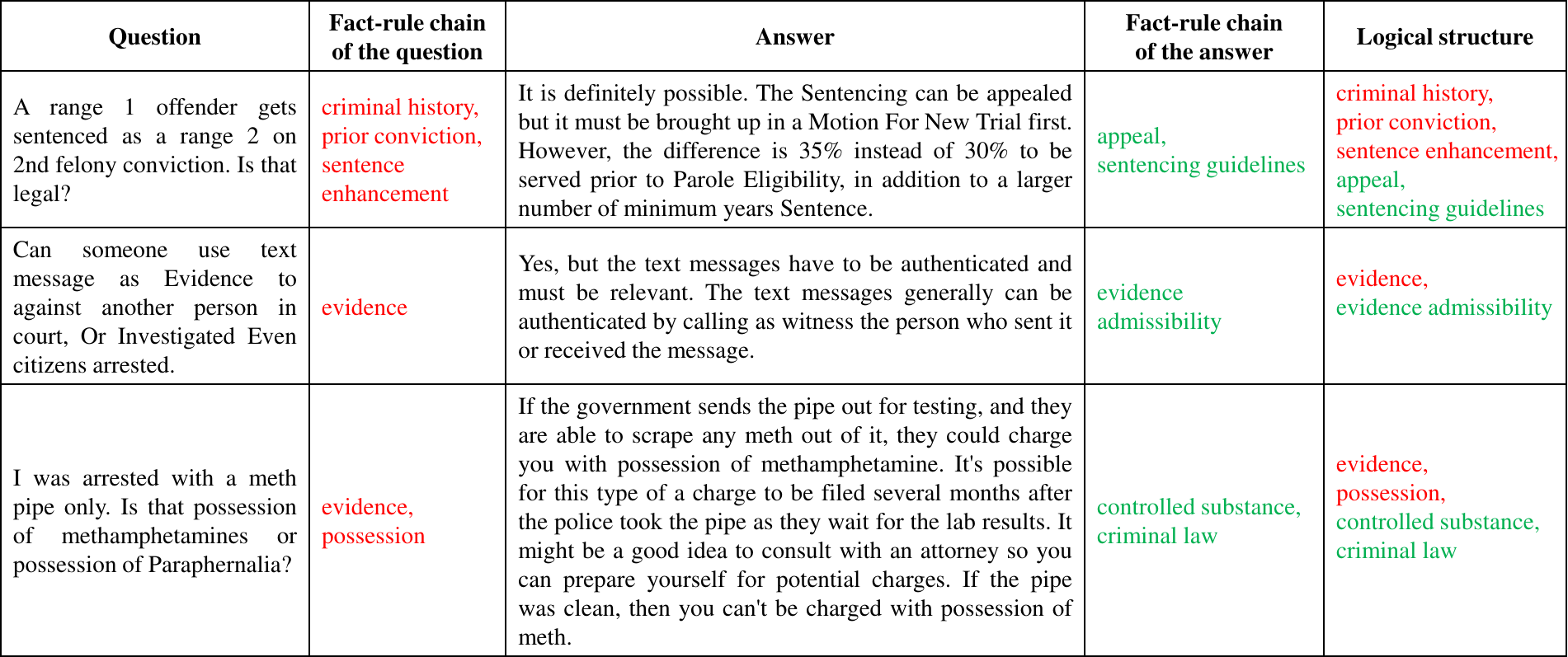}
    \caption{Illustrative examples of the fact-rule chains. The red text represents the fact-rule chain for the question, and the green text represents the fact-rule chain for the answer. Together, they form the logical structure.}
    \label{fig:fig6}
\end{figure*}

Illustrative examples of the fact-rule chains are provided in Figure~\ref{fig:fig6}.  

\section{Illustrative samples of the collected data}
\label{sec:appendix2}
Illustrative samples of the collected data are provided in Table~\ref{tab:appendixsample}.  

\begin{table*}[htbp]
\centering
\resizebox{\textwidth}{!}{%
\begin{tabular}{ccp{8cm}p{8cm}}
\toprule
ID & Location & User's question & Lawyer's answer \\
\midrule
1 & California & My boyfriend is incarcerated for PC 212.5, in 2017 he was sentenced to 9 years, 667a enhancements added to the sentencing. He has a parole release date in 2023, now after taking many classes and being in fire camp. Is there anyway to reduce more time after new laws passed regarding extra time for enhancements? & Prison prior-667.5(b) enhancements are now gone- but not 667a enhancements. If he successfully participated as a hand crew member at fire camp, once he's been released from custody, he can apply for relief from the convictions that sent him there, and parole, under the new 1203.4b P.C. \\
2 & Florida & My friend legally purchased but did not register a gun and it was found by police in the room she and her Bf share. He is a felon and took the charge because she has a pending DCF case. If she goes back and takes the charge, having no priors, is it possible she would just get probation and not jail time? & What would the charge be, failure to register a firearm? To my knowledge, Florida doesn't require registration, the boyfriend could be charged with being a felon in possession, but I don't think there is any charge she could "take". \\
3 & Florida & I requested an Officer to press charges for my stepson kicking me in the back and vandalizing my home with a sharpie. However, the officer said I did not adopt him and he is allowed to throw things around the house. I am trying to get a hold of a supervisor and I was told to call back when he is back from vacation. Please Advise. I feel strongly the police are bias and not doing there job. Also, my wife is trying to help me with full support of charges. This is odd of the Police. & So, it is definitely not the law that someone is allowed to kick you in the back and vandalize your house just because you didn't adopt him. The crime for vandalizing is called "Criminal Mischief," governed by Florida Statute 806.13. The crime for kicking you in the back is called "Battery," governed by Florida Statute 784.03. Neither crime has a defense or exception having to do with someone's adoption status. Good luck. \\
\bottomrule
\end{tabular}}
\caption{Illustrative samples of the collected data.}
\label{tab:appendixsample}
\end{table*}

\end{document}